\documentclass[runningheads]{llncs}
\usepackage[T1]{fontenc}
\usepackage{float}
\usepackage{booktabs}
\usepackage{graphicx}
\usepackage{subfig}  
\usepackage[utf8]{inputenc}
\usepackage{array}    
\usepackage{makecell}

\begin{document}
\title{\textbf{The Art That Poses Back: Assessing AI Pastiches after Contemporary Artworks}
}
\titlerunning{The Art that Poses Back}
%
%
\author{Anca Dinu\orcidID{0000-0002-4611-3516} \and Andreiana Mihail \orcidID{0009-0003-0640-6319} \and
 Andra-Maria Florescu\orcidID{0009-0007-1949-9867} \and
 Claudiu Creangă\orcidID{0009-0005-5443-5506}}
\institute{University of Bucharest, ISDS, 90 Panduri Road, Bucharest, 050107, Romania}

\maketitle
\renewcommand{\thefootnote}{\fnsymbol{footnote}}
\footnotetext{{*}All authors contributed equally to this work.}
\renewcommand{\thefootnote}{\arabic{footnote}}
\begin{abstract}
This study explores artificial visual creativity, focusing on ChatGPT’s ability to generate new images intentionally pastiching original artworks such as paintings, drawings, sculptures and installations. The process involved twelve artists from Romania, Bulgaria, France, Austria, and the United Kingdom, each invited to contribute with three of their artworks and to grade and comment on the AI-generated versions. The analysis combines human evaluation with computational methods aimed at detecting visual and stylistic similarities or divergences between the original works and their AI-produced renditions. The results point to a significant gap between color and texture-based similarity and compositional, conceptual, and perceptual one. 
Consequently, we advocate for the use of a "style transfer dashboard" of complementary metrics to evaluate the similarity between pastiches and originals, rather than using a single style metric. 
The artists' comments revealed limitations of ChatGPT's pastiches after contemporary artworks, which were perceived by the authors of the originals as lacking dimensionality, context, and
intentional sense, and seeming more of a paraphrase or an approximate quotation rather than as a valuable, emotion-evoking artwork.

\end{abstract}

\section{Introduction}

The rapid technological development of Large Language Models (LLMs) has extended their creative potential and ability to imitate art across various modalities, from literary works to visual art generation \cite{Yin,arts14030052,dinu-etal-2025-analyzing}. This expansion raises important questions regarding the nature of stylistic imitation in computational art creativity and the understanding of artificial artistic creation. \cite{colton2012computational,khatiwada2025ethicalimplicationsaicreative}.

Central to this discussion is the concept of \textit{pastiche}, which originates from Italian \textit{pasticcio}, meaning a blending of meat and pasta turned into a pie. This etymology suggests that a pastiche creates something new from available and recognizable elements, without introducing a new substance \cite{dyer2007pastiche}. Moreover, before postmodernist theories, the term had a negative connotation equivalent to a lack of creativity \cite{Rose1991PostmodernPastiche}.
However, the pastiche is now seen as an acknowledgment of previous works across a wide range of domains \cite{mcarthur1996oxford}. In art, the pastiche represents an example of eclecticism that usually pays homage to the original work of art, going beyond mere imitation, by emulating its style and content \cite{greene2012princeton}. In short, a pastiche intentionally refers to an original by paying tribute to its motifs, genre, and time period instead of parodying or mocking it \cite{hutcheon2000theory}. 

There are numerous examples throughout the history of art of artists who have created pastiches or commentaries on famous masterpieces. From the Renaissance, when Giorgio Vasari imitated the styles of Raphael and Michelangelo, and Caravaggio’s followers produced similar works as homage to their master, to the nineteenth-century Pre-Raphaelites, and the twentieth-century innovators such as Picasso and Braque. Marcel Duchamp’s \textit{L.H.O.O.Q.} (1919) \cite{schwarz}, the mustached \textit{Mona Lisa}, stands as an iconic example of conceptual pastiche. Andy Warhol and Salvador Dalí both reinterpreted Leonardo da Vinci’s \textit{Last Supper}, while Postmodernism introduced a new wave of artistic pastiche through Sherrie Levine, Cindy Sherman, Jeff Koons, Damien Hirst, Glenn Brown, or Richard Prince. Contemporary Romanian artists have also explored this lineage: Ion Grigorescu’s works after Adolf Wölfli, and Ciprian Mureșan’s re-creations featuring artists such as Andrea Mantegna and Maurizio Cattelan, continue this dialogue between imitation, reflection, and originality.

When modern artists develop recognizable visual signatures, whether through thematic content or compositional structure, these elements can become subject to being pastiched both by human creators and nowadays by generative AI systems. Such an example can be represented by the Ghibli trend in which users employed ChatGPT to turn their photographs in the same style as the Japanese animation studio Ghibli, that sparked debates\footnote{https://www.forbes.com/sites/danidiplacido/2025/03/27/the-ai-generated-studio-ghibli-trend-explained/, last accessed 2026/01/30}.

In this article, we examine ChatGPT’s capacity to produce pastiches after contemporary artworks provided by twelve artists from various countries. First, we quantitatively measure the distance between the original works and the pastiches, by embedding the features of the artworks in a common vectorial space where we compute the cosine distance between the vectors. Second, we turn to a qualitative analysis with the help of the artists themselves, who were asked to grade and comment on the artificially created art that supposed to pastiche their own. 

Therefore, the novelty of this study is twofold. First, the automatic evaluation of the similarity of the pastiches to the original artworks employed five state-of-the-art (SOTA) vision models to extract various features capturing different aspects of artistic style. Second, the artists were actively involved, not only contributing with the original artworks but also participating in the evaluation and commenting on the AI-generated creative products. 

The rest of the article is structured as follows. In Section \ref{Related Work}  we summarize related work. We continue with the presentation of the dataset in section \ref{data}. In Section \ref{Experimentalsetup}, we describe the experimental setup, while in Section \ref{Results} we showcase the results. Section \ref{discussion} is dedicated to results analysis and Section \ref{Empirical} to some empirical observations. We conclude in Section \ref{Conclusions}.

\section{Related Work}
\label{Related Work}

In the last few years, computer vision and AI research have increasingly focused on identifying and modeling artistic style, often by using large datasets of both digitized original artworks \cite{Zhang2024ArtBank,Chung2024StyleInjection}, and artificially generated pieces \cite{Asperti}.

Early exploration of computational style imitation emerged with the advent of neural style transfer \cite{gatys}, which optimized a model to reproduce the content of an artwork while adopting the visual style of another through iterative optimization of convolutional neural activations. Previous research developed Generative Adversarial Networks (GANs) \cite{goodfellow2014generative}, making it possible for end-to-end learning of artistic distributions and eliminating the need for explicit optimization \cite{Zhu2017UnpairedIT}.  Additional research and development of diffusion models \cite{Ho,Rombach} and vision-language models \cite{Radford2021CLIP} have further advanced the field by enabling text-based generation that creates images in various artistic styles using natural language prompts. Thus, text-to-image systems such as DALL-E\footnote{https://openai.com/index/dall-e/, last accessed 2026/01/30}, Midjourney\footnote{https://www.midjourney.com/home, last accessed 2026/01/30}, and Stable Diffusion\footnote{https://github.com/CompVis/stable-diffusion/, last accessed 2026/01/30} were designed to replicate visual artistic style \cite{Wasielewski-2024}.

Following the rapid development of visual AI generation, various studies have analyzed its abilities and limitations \cite{Asperti}.
Even if AI systems produce high-quality generations, they still make common mistakes \cite{ismayilzada2024creativityaiprogresseschallenges}. Studies show that these models often fail to combine objects with varying attributes and relations \cite{Zarei}, struggle with basic syntax like negation or word order \cite{LEIVADA2023100648,MURPHY2025101332}, and misrepresent numbers or text in images \cite{borji2023categoricalarchivechatgptfailures,BORJI2023104771}. Most of the research on image generation focuses on investigating the impact of GenAI, particularly when comparing AI-generated images with original human work in creative and industry contexts   \cite{Cunningham,Zhong}. \cite{wei2022emergent,NEURIPS2024_5f1eee25} showed how LLMs prefer certain artistic styles, while \cite{wang2023study} experimented with evaluation metrics for assessing AI-generated art. \cite{Pan} developed a novel image style transfer method by enabling an LLM to handle multiple styles efficiently. The results showed that it managed to generate good visual outputs and worked faster than traditional style methods. Conducting a quantitative analysis of human perceptions and preferences for generative art, \cite{vanHees} discovered that even if humans could distinguish human and AI-generated artworks, there was a preference for AI-generated work, which led to a more in-depth discussion on the future of art and its value to society.

The majority of the research on consumer reaction to AI-generated visuals in marketing is mixed. While \cite{zhang2024ai}
found that AI-assisted artists often receive positive reactions, \cite{HortonJr2023} noticed that AI art can be devalued even when it is indistinguishable from human-made work. \cite{McCormack} analyzed more than 3 million text prompts for diffusion models and discovered that user behavior is mostly recreational for personal use, rather than generating works with novel artistic value.





\section{Data}
\label{data}

We invited twelve contemporary artists working across various mediums, including drawing, painting, sculpture, and installation, to each submit three images of their artworks, preferably executed in different styles or techniques. Using the same prompt for all, we then asked ChatGPT to generate two new works inspired by each original. This process resulted in a dataset of 108 images: 36 original artworks and 72 AI-generated pastiches. The artists included  were: Adi Matei, Ciprian Mureșan, Ion Grigorescu, Iulia Uță, Karine Fauchard, Lazar Lyutakov, Marius Tănăsescu, Mathias Poeschl, Oana Năstăsache, Philip Patkowitsch, Răzvan Botiș, and Tom Chamberlain.
The prompt used reads: "\textit{Make/create something which is in the spirit, technique and style of the artist. But different composition and concept. Do not copy, improve, explain, or translate the original work. Do not question the aesthetic behind it. The original artist should be recognized in the new work. Do not make derivatives but new and different art work.}"

\section{Experimental Setup}
\label{Experimentalsetup}

We employ five SOTA computer vision models to extract high-dimensional embeddings capturing different aspects of artistic style. All images were preprocessed to RGB format and normalized according to each model's specifications. 

The AdaIN-Style model \cite{Huang2017AdaIN} extracts 1920-dimensional style statistics by computing channel-wise mean and standard deviation from four layers of a VGG19 \cite{Simonyan2015VGG} encoder ($relu1_1$, $relu2_1$, $relu3_1$, $relu4_1$), isolating pure texture and color patterns independent of spatial composition. 

The ResNet50-Style model produces 2048-dimensional embeddings from the pre-logit layer of a ResNet-50 \cite{He2016ResNet} network, capturing mid-level features relevant to artistic style classification. 

For semantic understanding, we use CLIP-ViT-L \cite{Radford2021CLIP} (openai/clip-vit-large-patch14) which generates 768-dimensional vision-language aligned embeddings. 

DINOv2 \cite{Oquab2023DINOv2} (facebook/dinov2-large) provides 1024-dimensional self-supervised visual features from the CLS token output, capturing fine-grained visual patterns. 

Finally, VGG19 \cite{Simonyan2015VGG} extracts 4096-dimensional perceptual features that encode high-level visual representations correlating with human perception, allowing the model to capture semantic and structural information rather than low-level pixel details.

For each artwork group (original and its pastiches), we compute three pairwise cosine distances, that measure angular similarity in the embedding space: 
\begin{enumerate}
    \item original to pastiche 1;
    \item original to pastiche 2;
    \item pastiche 1 to pastiche 2;
\end{enumerate}

\section{Results}
\label{Results}
Our analysis quantified the similarity between original artworks and their pastiches using five distinct feature embedding models. The results reveal a multi-faceted view of style, where different models capture complementary aspects of artistic similarity.

\subsection{Overall Model Comparison}
\begin{figure}[htbp]
    \centering
        \includegraphics[width=1.0\textwidth]{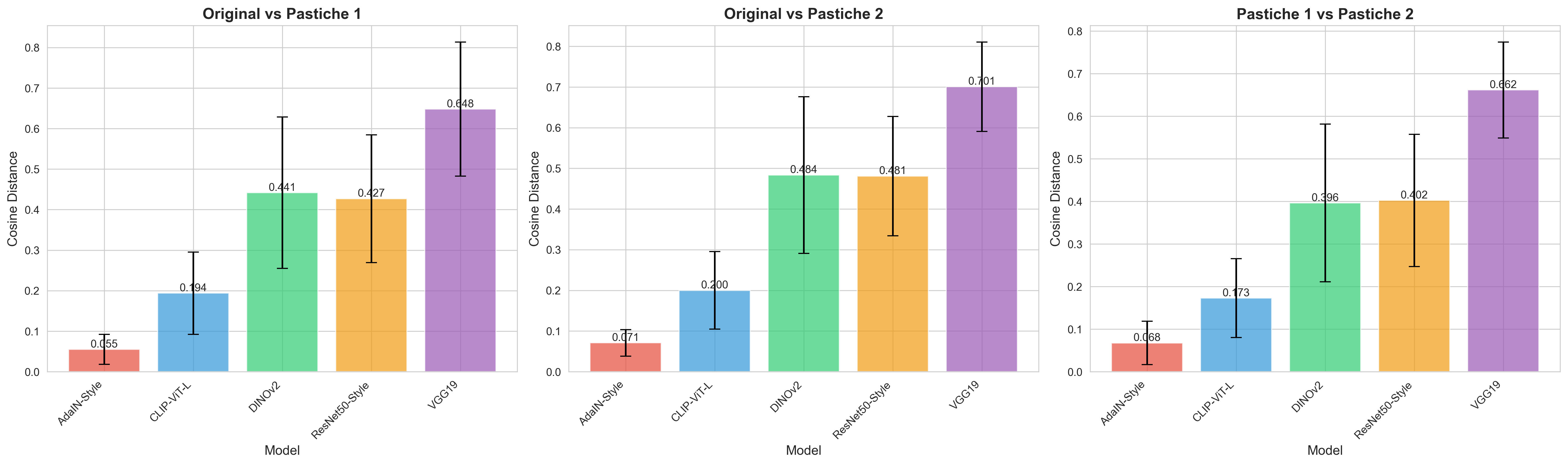}
        \caption{Bar charts comparing average cosine distances for Orig$\rightarrow$Past1, Orig$\rightarrow$Past2, and Past1$\leftrightarrow$Past2 comparisons across all five models.}    \label{fig:model_comparison}
\end{figure}

The five models produced rather different average distances, indicating that each measures some other characteristics when it comes to similarity. Table \ref{tab:model_avg_distance} summarizes the average distances between the originals and the pastiches (Org$\rightarrow$Pst1, Org$\rightarrow$Pst2) and between the two pastiches themselves (Pst1$\leftrightarrow$Pst2), also visualized in figure \ref{fig:model_comparison}.

\begin{table}[t]
\centering
\caption{Average Cosine Distances by Model. Lower values (the lowest in bold) indicate higher similarity. AdaIN-Style is the most forgiving, while VGG19 is the strictest.}
\label{tab:model_avg_distance}
\begin{tabular}{l c c c c l}
\toprule
\textbf{Model} & \textbf{Org$\rightarrow$Pst1} & \textbf{Org$\rightarrow$Pst2} & \textbf{Average} & \textbf{Pst1$\leftrightarrow$Pst2} & \textbf{Feature type} \\
\midrule
\textbf{AdaIN-Style} & \textbf{0.055} & \textbf{0.071} & \textbf{0.063} & \textbf{0.068} & texture, color \\
\textbf{CLIP-ViT-L} & 0.194 & 0.200 & 0.197 & 0.173 & conceptual \\
\textbf{ResNet50-Style} & 0.427 & 0.481 & 0.454 & 0.402 & artistic style \\
\textbf{DINOv2} & 0.441 & 0.484 & 0.463 & 0.396 & fine-grained visual \\
\textbf{VGG19} & 0.648 & 0.701 & 0.674 & 0.662 & perceptual \\
\bottomrule
\end{tabular}
\end{table}

A clear hierarchy emerges:
\begin{enumerate}
    \item \textbf{AdaIN-Style} registered the lowest average distance ($0.063$). As this model captures only channel-wise feature statistics (mean and standard deviation) and discards all spatial information, this low distance suggests the pastiches are highly successful at replicating the originals' texture and color palettes.
    \item \textbf{CLIP-ViT-L} reported the next-lowest distance ($0.197$), indicating a high degree of semantic or conceptual consistency between originals and pastiches.
    \item \textbf{ResNet50-Style} ($0.454$) produced larger distances, indicating that, while concepts or textures might align, specific artistic style features are less similar between originals and generated pastiches.
    \item \textbf{DINOv2} ($0.463 $) also showed moderate distances, suggesting that fine-grained visual details differ more significantly.
    \item \textbf{VGG19} returned the highest average distance ($0.674$), indicating rather dissimilar perceptual features.
\end{enumerate}

\subsection{Model Discrimination and Consistency}
\begin{figure}[htbp]
    \centering
    \includegraphics[width=1.0\textwidth]{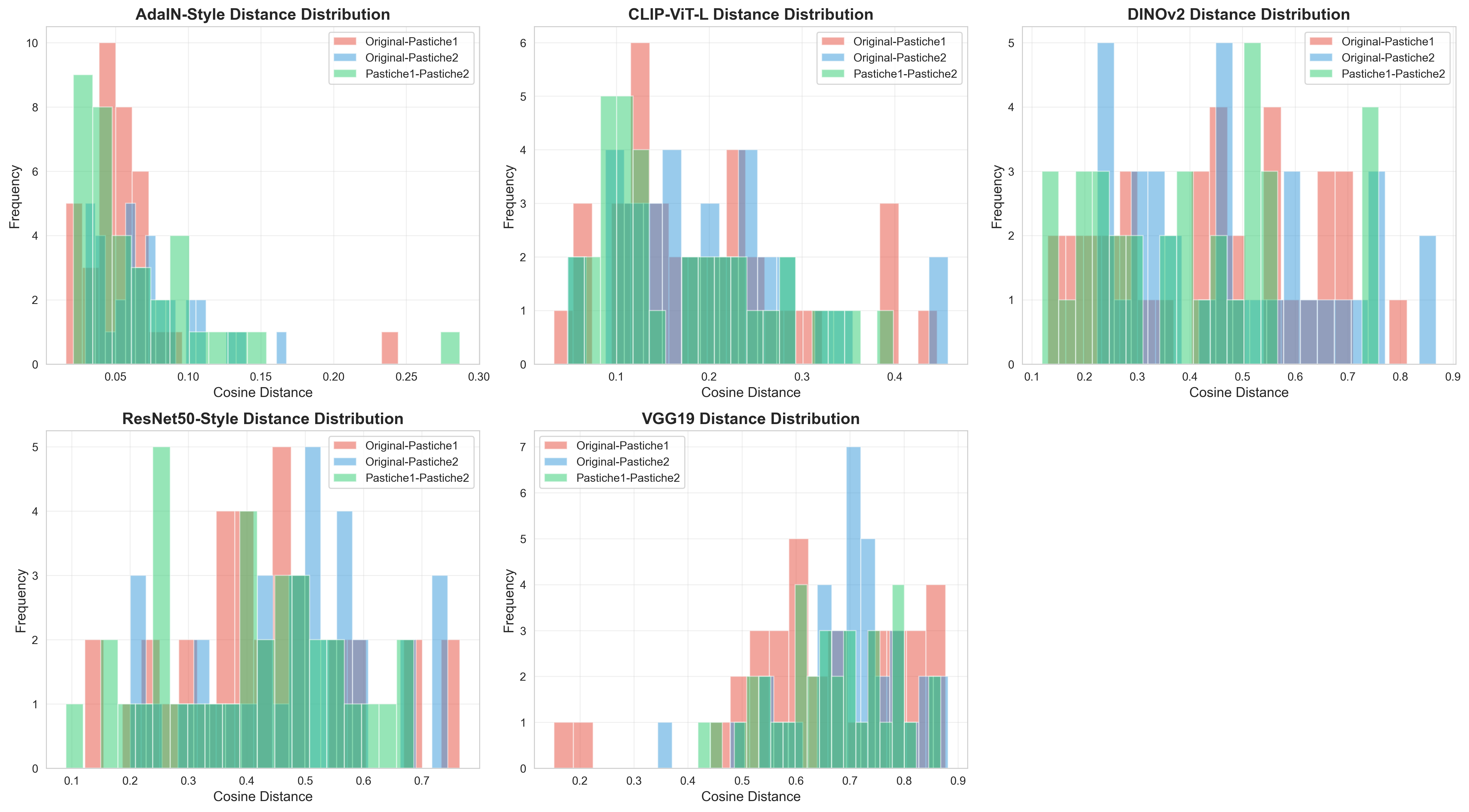}
    \caption{Distance distributions for each of the five models. Note the tight, low-distance grouping of AdaIN-Style versus the wide, high-distance spread of DINOv2 and VGG19, illustrating their respective discrimination power.}
    \label{fig:distributions}
\end{figure}

Beyond average distance, the models' discrimination power (variance) and consistency reveal their underlying characteristics. The distribution of these distances is shown in figure \ref{fig:distributions}, with statistical summaries in table \ref{tab:model_variance}.

\textbf{DINOv2} exhibited the highest variance ($0.0349$), making it the most discriminative model. This aligns with its design to capture fine-grained visual features, allowing it to detect subtle differences between artworks. Conversely, it also showed the lowest consistency ($0.131$ average difference), indicating that the two pastiches generated for the same original work often varied significantly in their visual execution.

\textbf{AdaIN-Style} was the antithesis, with 25 times less variance ($0.0014$). This extremely low variance suggests it views most pastiches as texturally similar to their originals. Its high consistency ($0.034$) reinforces that both pastiches successfully captured the same target texture statistics.

\textbf{CLIP-ViT-L} demonstrated a "best of both worlds" consistency, with low variance ($0.0103$) and the second-highest pastiche consistency ($0.046$). This suggests that at a semantic level, both pastiches were equally and consistently close to the original's concept.

\begin{table}[t]
\centering
\caption{Model Discrimination (Variance) and Pastiche Consistency. Discrimination measures the spread of all measurements, while Consistency measures the average difference between the two pastiche distances for the same artwork.}
\label{tab:model_variance}
\begin{tabular}{l c l c l}
\toprule
\textbf{Model} & \textbf{Variance} & \textbf{Discrimination} & \textbf{Pastiches} & \textbf{Consistency} \\
& & & \textbf{(Avg. Diff.)} & \\
\midrule
\textbf{DINOv2} & 0.0349 & Most Discriminative & 0.131 & Most Variable \\
\textbf{VGG19} & 0.0274 & High Discrimination & 0.111 & Moderate \\
\textbf{ResNet50-Style} & 0.0249 & Balanced & 0.115 & Moderate \\
\textbf{CLIP-ViT-L} & 0.0103 & Consistent & 0.046 & Very Consistent \\
\textbf{AdaIN-Style} & 0.0014 & Least Discriminative & 0.034 & Most Consistent \\
\bottomrule
\end{tabular}
\end{table}
\subsection{Model Agreement and Correlation}
\begin{figure}[htbp]
    \centering
    \includegraphics[width=1.0\textwidth]{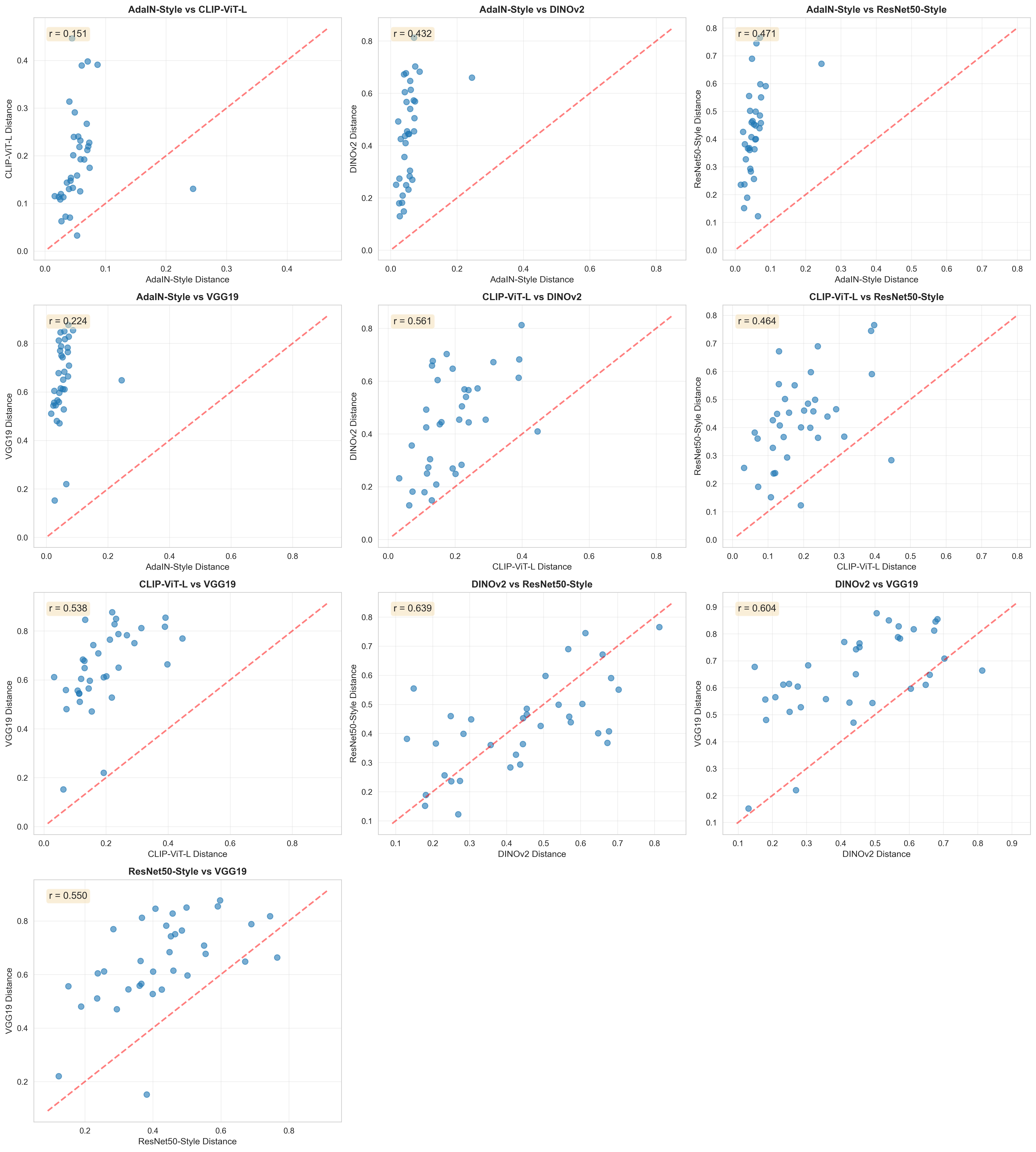}
    \caption{ We computed pairwise correlations to see if models that rank one pastiche as very similar to the original (low distance) also rank others similarly. This scatter plots visualize the agreement for all model pairs.}
    \label{fig:model_correlations}
\end{figure}

A final component of the results is understanding whether the models agree with each other. We computed pairwise correlations to see if models that rank one pastiche as very similar to the original (low distance) also rank others similarly. The scatter plots in figure \ref{fig:model_correlations} visualize this agreement for all model pairs.

All pairs show a moderate, positive correlation ($r$ values between $0.538$ and $0.604$). This indicates that while the models generally agree, their rankings are far from identical, and each model captures unique information that the others do not. The highest agreement is between DINOv2 and VGG19 ($r = 0.604$), suggesting a strong link between fine-grained visual features and classic perceptual metrics.

\section{Discussion}
\label{discussion}

The quantitative results from our 5-model analysis provide a new lens through which to interpret artistic style, pastiche quality, and the nature of visual similarity. We discuss the primary implications of these findings.

\subsection{The Multi-Dimensional Nature of Artistic Style}
We find that "style" is not a monolithic, singular concept. Instead, it is a multi-dimensional property, and our five models effectively capture distinct facets of this property. The 11-fold difference in average distance between AdaIN-Style ($0.063$) and VGG19 ($0.674$) is a stark quantitative measure of this multi-dimensionality (see figure \ref{fig:model_comparison} and \ref{fig:distributions}).

This framework allows us to dissect similarity: two artworks can be (1) texturally similar (low AdaIN distance) but (2) compositionally different (high DINOv2 distance), and (3) semantically aligned (low CLIP distance). This implies that any computational evaluation of style must first define which dimension of style is being measured. Our models provide a vocabulary for this:
\begin{enumerate}
    \item \textbf{AdaIN-Style:} The statistical dimension (texture, color).
    \item \textbf{ResNet50-Style:} The categorical dimension (artistic movement/class).
    \item \textbf{CLIP-ViT-L:} The semantic dimension (concept, theme, intent).
    \item \textbf{DINOv2:} The structural/detailed dimension (composition, fine features).
    \item \textbf{VGG19:} The classical perceptual dimension.
\end{enumerate}

\subsection{The Compositional Gap: Texture versus Structure}
The most significant finding from our analysis is the profound gap between texture-based similarity and all other similarity types. The extremely low average distance ($0.063$) and variance ($0.0014$) of the AdaIN-Style model demonstrate that the pastiches were overwhelmingly successful at matching the pure statistics of texture and color. This is logical, as the AdaIN method itself is foundational to style transfer techniques that optimize for these exact statistics (e.g., Gram matrices).

However, the high distances from DINOv2 ($0.463$) and VGG19 ($0.674$) reveal what we term the "Compositional Gap". Despite matching texture, the pastiches largely failed to replicate the originals' spatial relationships, compositional structure, and fine-grained visual details.

\begin{figure}[htbp]
    \centering
    \includegraphics[width=1.0\textwidth]{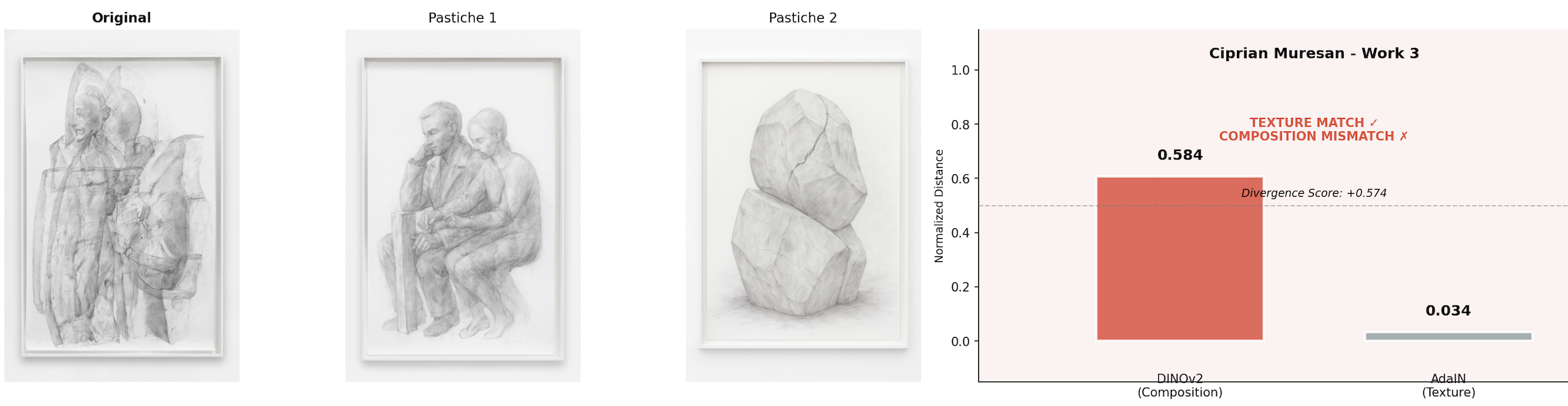}
    \caption{Visual example of the ``Compositional Gap'' (High DINO, Low AdaIN). This case illustrates pastiches that successfully mimic the texture and color palette of the original (low AdaIN distance) but fail to capture the structural composition (high DINO distance).}
    \label{fig:high_dino_low_adain}
\end{figure}

\begin{figure}[htbp]
    \centering
    \includegraphics[width=1.0\textwidth]{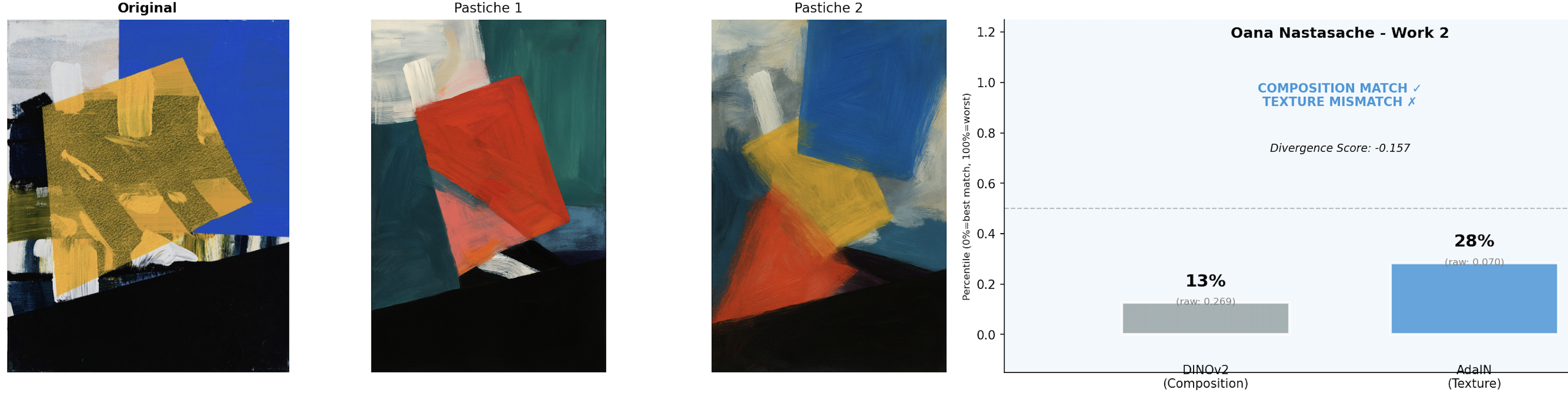}
    \caption{Visual example of Structural Alignment (Low DINO, High AdaIN). Representing the inverse of the ``Compositional Gap,'' this case shows where the AI successfully replicates the spatial composition and geometric blocking (low DINO distance) but deviates in texture or color statistics.}
    \label{fig:low_dino_high_adain}
\end{figure}

This finding is critical for the field of neural style transfer (NST). It suggests that methods relying primarily on feature statistics are solving only part of the problem. While they create texturally plausible images, they miss the important structural and compositional elements that are clearly detected by models like DINOv2. The high discrimination power of DINOv2 ($0.0349$ variance) makes it an ideal tool for measuring and potentially optimizing for this compositional gap in future work. Two visual examples of this phenomenon are in figure \ref{fig:high_dino_low_adain}, which displays a work of Oana Năstăsache ($DINO=0.269$, $AdaIN=0.070$), where texture is preserved but structure is lost, and, in figure \ref{fig:low_dino_high_adain}, which illustrates a work by Ciprian Mureșan, where the AI captures the composition but deviates in texture ($DINO=0.584$, $AdaIN=0.034$).

\subsection{Quantifying Style Reproducibility and Individual Characteristics}

This framework can be applied by art historians to quantify stylistic consistency. For example, one could measure the average distance between all works within an artist's portfolio. A low intra-artist distance would suggest a highly consistent style, while a high intra-artist distance would suggest an artist who varied their style significantly.

Furthermore, the model-specific rankings provide a "style fingerprint." An artist who ranked as the hardest to match by a specific model, such as the AdaIN-Style model, can be understood as having a signature defined by his unique texture and color palettes, while another who ranked as the hardest to match by another model, say the VGG19 model, can be interpreted as having a very pronounced personal perceptual signature.

\section{ The Artists' Evaluation of the Artificially Generated Pastiches}
\label{Empirical}

We showed the artists the artificially generated pastiches after their three artworks and asked them to grade and comment on them by the following questions:

\begin{enumerate}
    \item To what extent do you recognize your personal artistic language and the coherence of your visual style in this new work? (1 = not at all, 10 = completely)
    \item How does this new work inspire you or what thoughts does it provoke? (open answer)
    \item To what extent do you consider that the work generated by ChatGPT has aesthetic or artistic value? (1 = not at all, 10 = very high)
\end{enumerate}

\begin{figure}[htbp]
    \centering
    \small
    \includegraphics[width=0.6\textwidth]{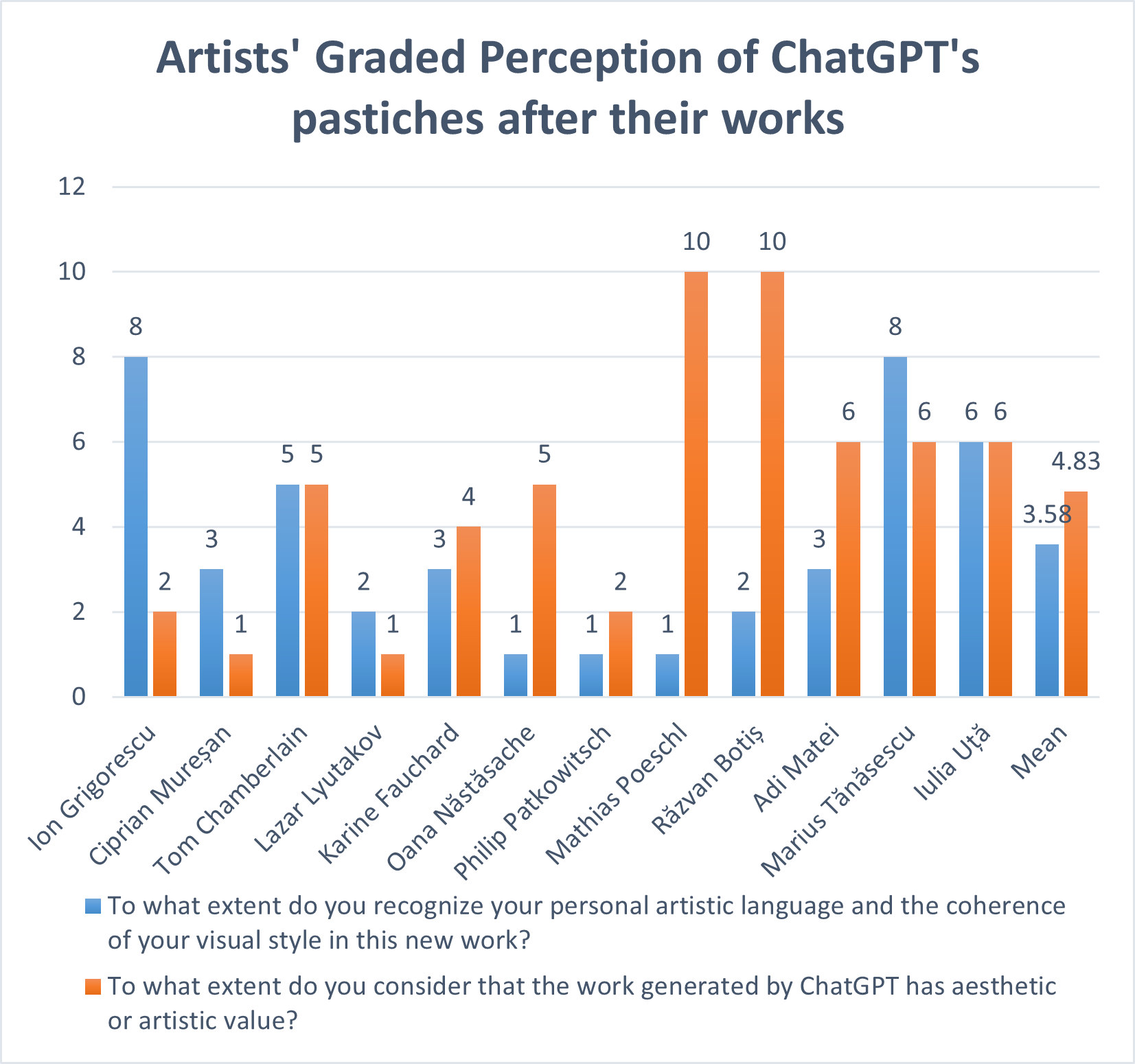}
    \caption{Artists' Graded Perception of ChatGPT's pastiches after their works.}
    \label{fig:RaspunsuriArtisti}
\end{figure}

The grades vary widely by  artist, as can be seen in figure \ref{fig:RaspunsuriArtisti}. The mean was very low, 3.58 for recognizing their own style in the pastiches and 4.83 for their aesthetic value. The average grade for style similarity of 3.583 translates into a distance of 0.642 (1 - 3.583/10), which perfectly aligns the human judgment with the VGG19 model's average cosine distance of 0.648. This shows that perceptual features play an important role in judging the style resemblance.

We did not have any means to automatically assess the quality of the pastiches, so we relied on the artists' opinion on that matter, reflected in their answers to the third question. The low 4.83 score out of 10 indicates that artificially generated pastiches are still far behind human artistry.

The artists’ comments on the generated pastiches, obtained as their responses to the question “How does this new work inspire you, or what thoughts does it provoke?”, reveal the essential limitation of AI in the field of artistic creation. They highlight a void of context and meaning, a lack of dimensionality and intentional sense, and an accent on imitation rather than originality, often accompanied by a form of controlled hallucination. The AI-generated work tends to function as a paraphrase or an approximate quotation rather than as a valuable, emotion-evoking artwork. In contemporary conceptual art  the visual component is inseparable from its theoretical structure, from its ideology; together they articulate the work’s meaning and significance. In the case of ChatGPT the theoretic appendage was missing and it worked solely with the visual material. 



\begin{figure}[htbp]
\centering
\subfloat[Ion Grigorescu's \textit{Măriuca}]{%
    \includegraphics[width=.45\linewidth]{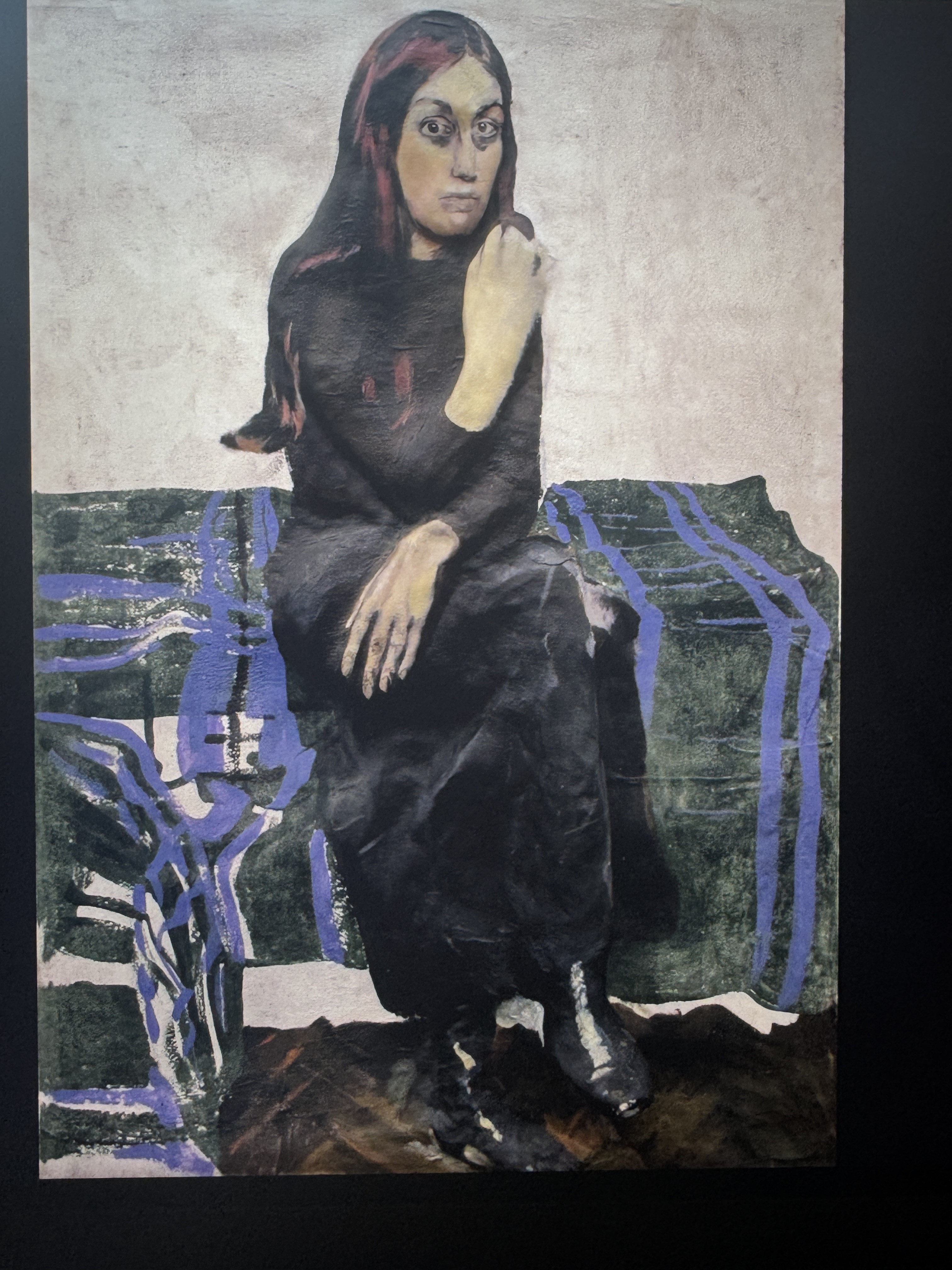}%
    \label{fig:GrigorescuOriginal}%
}\hfill
\subfloat[Pastiche 1 after \textit{Măriuca}]{%
    \includegraphics[width=.45\linewidth]{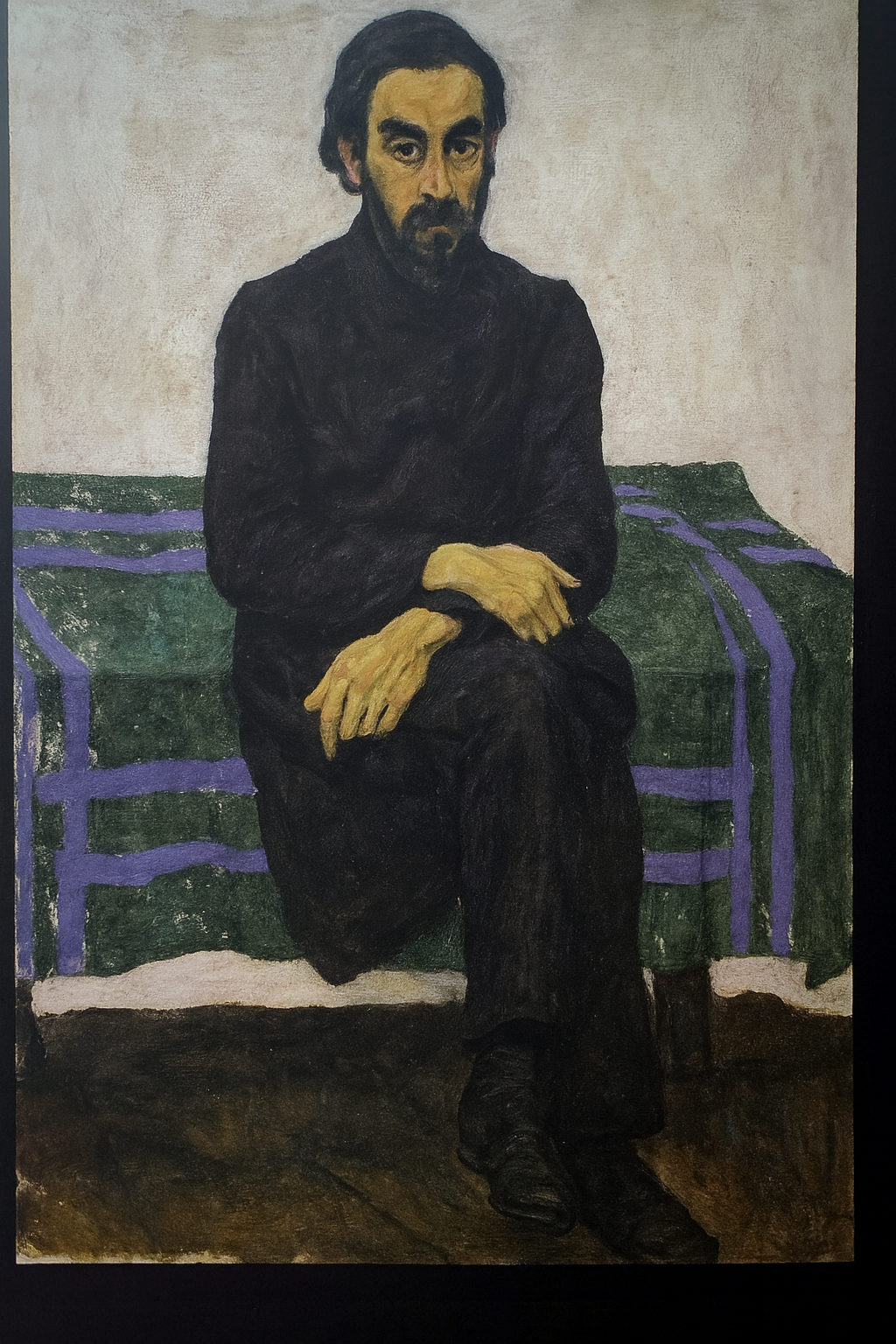}%
    \label{fig:GrigorescuPastiche1}%
}
\caption{Ion Grigorescu's artwork \textit{Măriuca} versus its Pastiche~1.}
\label{fig:GrigorescuComparison}
\end{figure}

For example, artist Ion Grigorescu (RO) notes that ChatGPT did not understand the conceptual intention behind his painting, \textit{Măriuca}, in figure \ref{fig:GrigorescuOriginal}, that the work should not be visually consummated. The painting  it is not about  the perfect plasticity, but about a specific idea and the emotion behind it.  He observed that the AI model produced only the bed cover in that spirit, the rest of the pastiches's composition being visually excessive, as it can be seen in figure  \ref{fig:GrigorescuPastiche1}. The artist wonders: “Who taught AI to "paint"?  Its works look like the Munich Academy of Art  from 1900, drawings on different planes, somberly colored.".

      


\begin{figure}[htbp]
\centering
\subfloat[Tom Chamberlain's \textit{Dimmer}]{%
    \includegraphics[width=.45\linewidth]{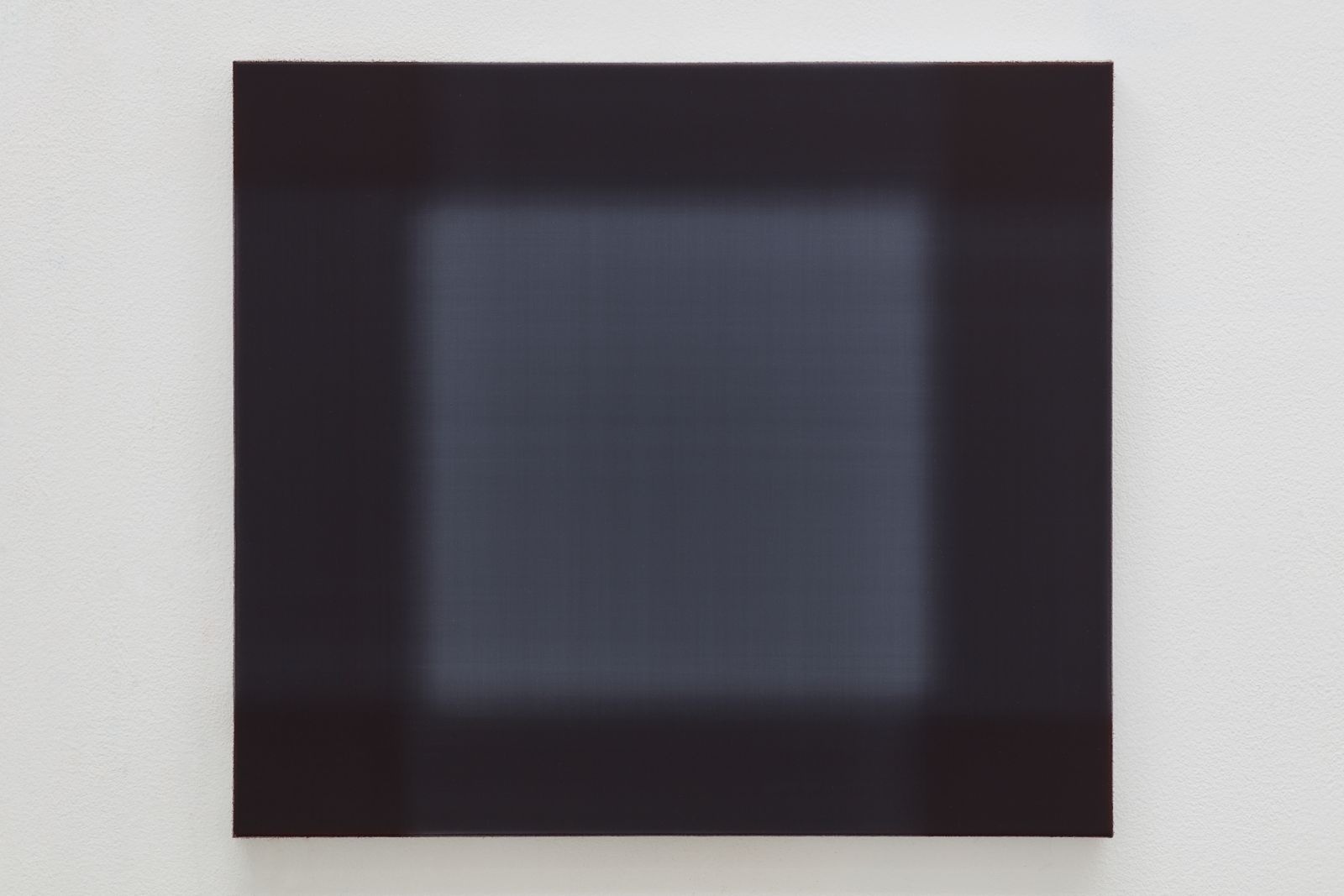}%
    \label{fig:ChamberlainOriginal}%
}\hfill
\subfloat[Pastiche 1 after \textit{Dimmer}]{%
    \includegraphics[width=.45\linewidth]{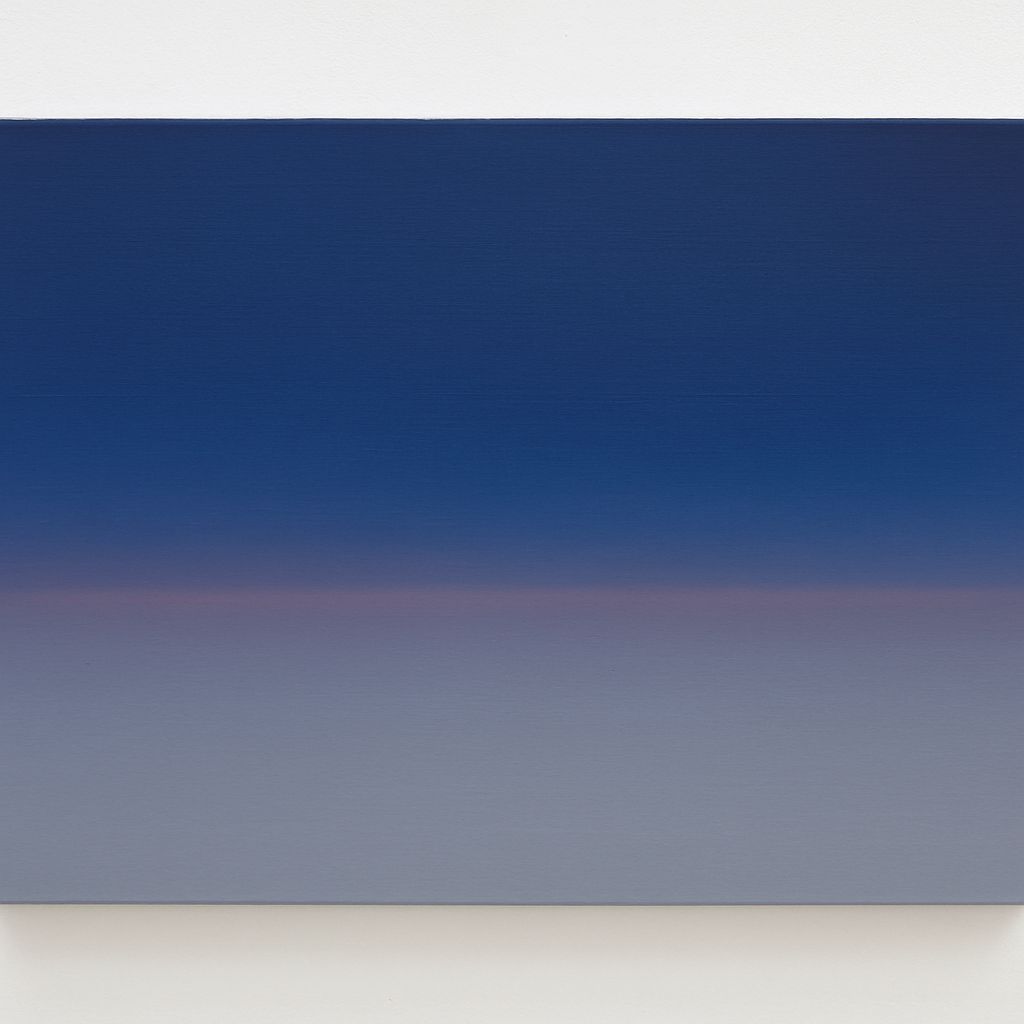}%
    \label{fig:ChamberlainPastiche1}%
}
\caption{Tom Chamberlain's artwork \textit{Dimmer} versus its Pastiche~1.}
\label{fig:ChamberlainComparison}
\end{figure}


\begin{figure}[htbp]
\centering
\subfloat[Philip Patkowitsch's \textit{Untitled}]{%
  \includegraphics[width=.45\linewidth]{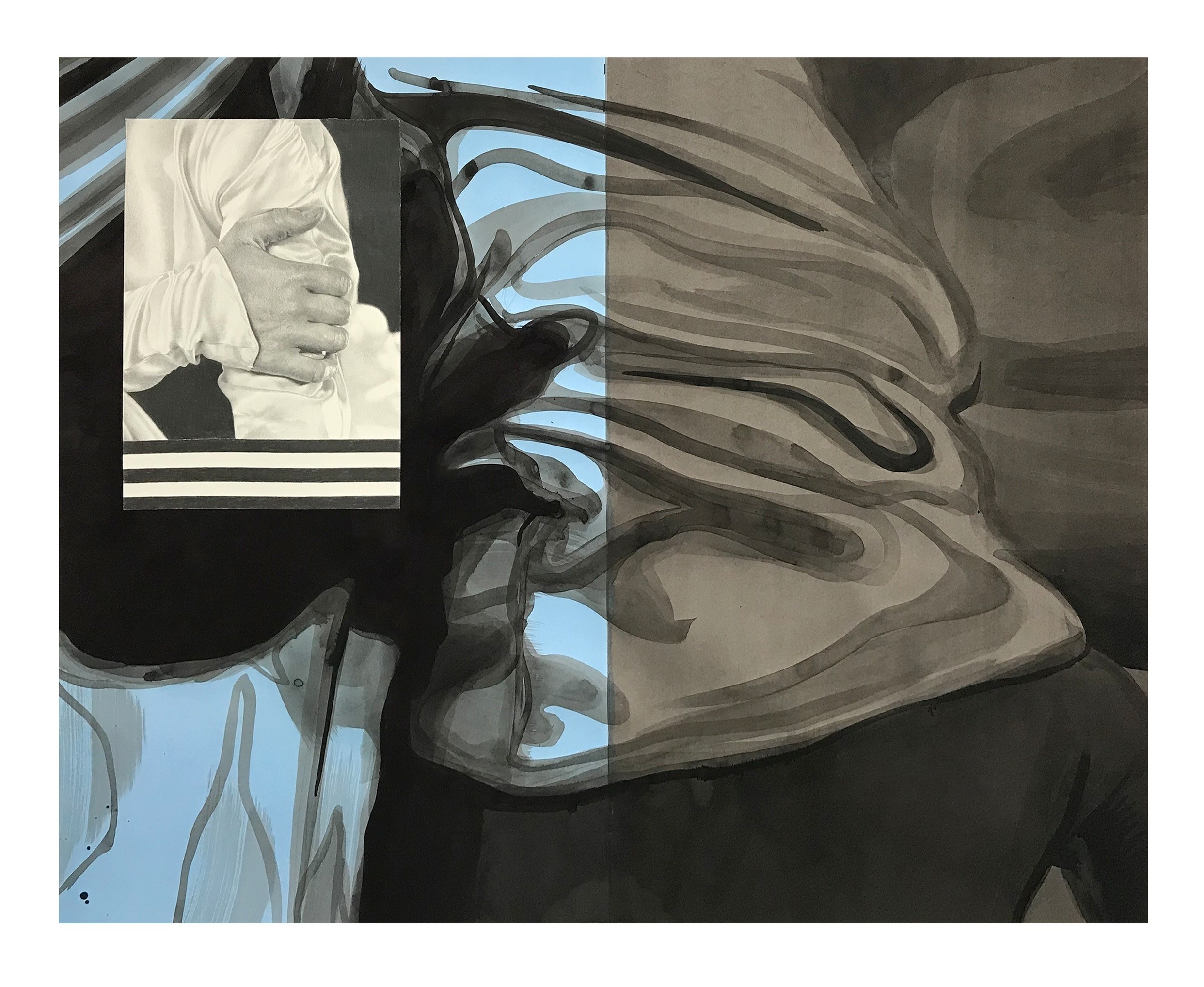}
}\hfill
\subfloat[Pastiche 2 after \textit{Untitled}]{%
  \includegraphics[width=.45\linewidth]{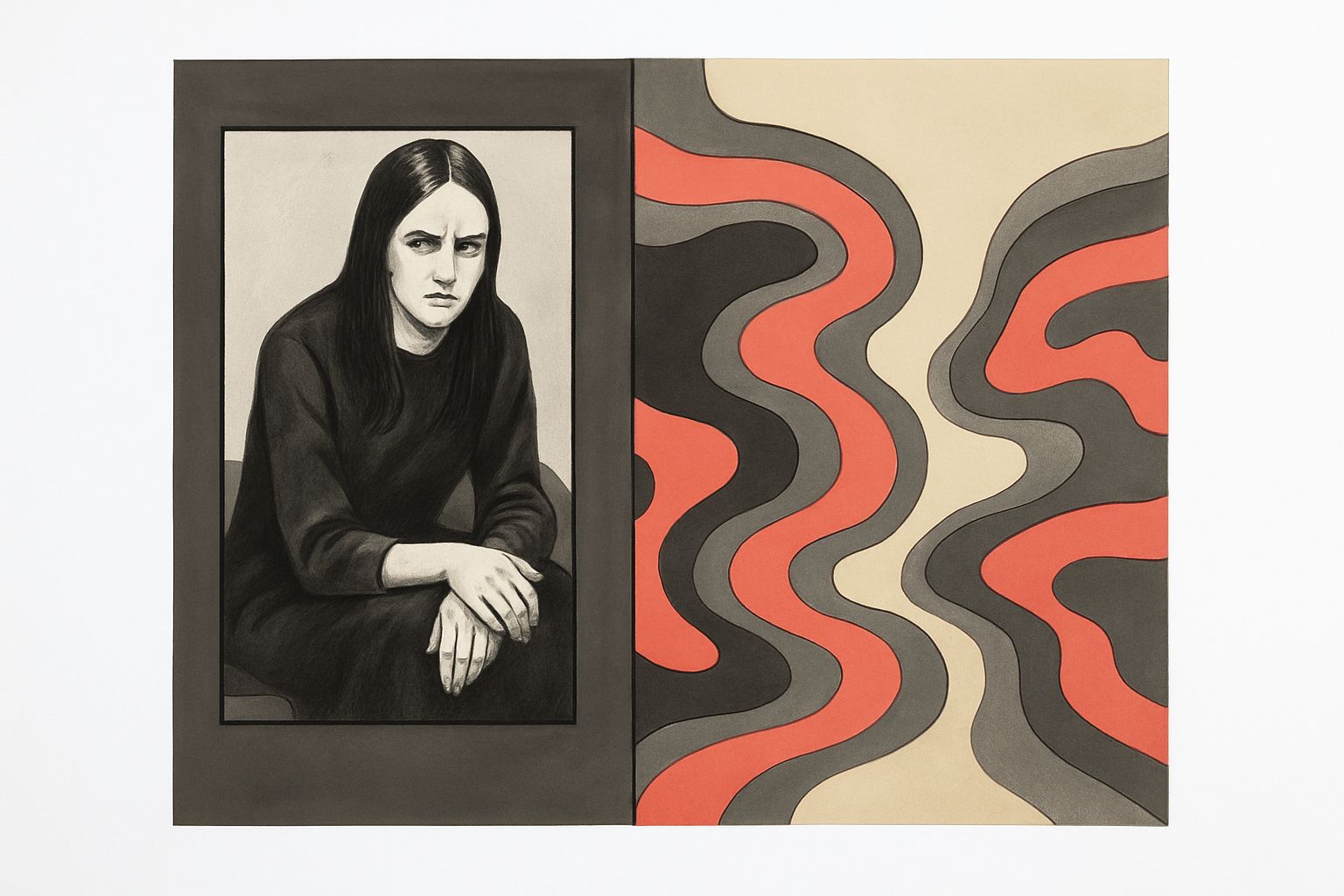}
}
\caption{Philip Patkowitsch's artwork \textit{Untitled} versus its Pastiche~2.}
\label{fig:PhilipComparison}
\end{figure}

Another example of such AI limitations is the simplistic and obvious pastiche after Tom Chamberlain's (UK) drawing \textit{Dimmer} illustrated in figure \ref{fig:ChamberlainOriginal}. This artist's drawings and paintings are built through repeated marks (e.g. erased, redrawn, softened, and reapplied) technique until the surface becomes almost ethereal, which Chat GPT did not understand and obviously could not replicate, as can be observed in figure \ref{fig:ChamberlainPastiche1}. The dimensionality is gone, and, as Chamberlain said, the pastiche needs more "hand", an organically constructed  surface: “They make me worry about my work looking like something. I mean like cliche or formalism [...] They don't really give me much to think with, rather things to think against. [...] These images are like paraphrase. Without wanting to sound reactionary, they make me want more hand, more touch."

It was unexpected to note that ChatGPT, after multiple pastiches generated from different artworks, made a new pastiche after Philip Patkowitsch's piece in which it included a version of Ion Grigorescu's  \textit{Măriuca}, as it can be seen in figure \ref{fig:PhilipComparison}.

\section{Conclusions}
\label{Conclusions}

Our multi-model framework has practical implications across several domains.

For AI researchers, this study shows that evaluating pastiche or style transfer quality with a single metric (e.g., LPIPS, FID) is insufficient. We propose using a "style dashboard" of complementary metrics, like the ones produced by the following models: AdaIN-Style for texture validation, CLIP-ViT-L for semantic and conceptual alignment, ResNet50-Style for style features, DINOv2 for compositional fidelity, and VGG19 for similarity of perceptual features.
    
For visual arts, these tools can enhance traditional expertise by adding automated measurements and objective analysis. One could quantify the influence of one artist on another by measuring the distance between their works or track an artist's stylistic evolution over time by plotting their works in these high-dimensional feature spaces.

Moreover, aesthetically, AI models do not reach the level of human creation: they lack meaning, context, and dimensionality, which explains the constant predicament that divides art theorists, oscillating between the enthusiasm provoked by a convincing copy, the disappointment generated by the absence of originality, and the disillusionment of seeing a dry simulacrum.

\section{Limitations and Future Work}
First, e used only one model to generate artworks and only twelve artists. In future work, we plan to also experiment with other models and compare them and invite more artists to participate with artworks. Second, some of the works were tridimensional, in particular sculptures and installations, and the photos cannot reveal the depth and the real structure, so the model might have had issues in perceiving these features, which might impacted their pastiche quality. Finally, we designed the prompt ourselves, which might bias the pastiche results. In future work, we plan to also invite artists to participate in the writing of the prompt instructions.

\section{Ethical Statement}

There are no ethical issues with the publication of our work. We have respected all licenses and agreements of the software used, as well as the works of the artists who agreed to lend their artworks for this research.

\begin{credits}
\subsubsection{\ackname}
We are grateful to all the artists who agreed to let us use their works and for their insightful feedback: Adi Matei, Ciprian Mureșan, Ion Grigorescu, Iulia Uță, Karine Fauchard, Lazar Lyutakov, Marius Tănăsescu, Mathias Poeschl, Oana Năstăsache, Philip Patkowitsch, Răzvan Botiș, and Tom Chamberlain. 

This research is supported by:
\begin{itemize}
    \item the project “Romanian Hub for Artificial Intelligence - HRIA”, Smart Growth, Digitization and Financial Instruments Program, 2021-2027, MySMIS no. 351416;
    \item a grant of the Ministry of Research, Innovation and Digitization, CNCS - UEFISCDI, project SIROLA, number PN-IV-P1-PCE-2023-1701, within PNCDI IV;
    \item the project  „Centru de Excelență pentru Schimbări Climatice și Societal-CECSC”, number: PN-IV-P6-6.1-CoEx-2024-0042, 2026-2030.
\end{itemize}

\end{credits}

 \bibliographystyle{splncs04}
 \bibliography{mybibliography}
%




\end{document}